\documentclass[11pt]{article}

\usepackage[final]{acl}

\usepackage{times}
\usepackage{latexsym}
\usepackage{orcidlink}

\usepackage[T1]{fontenc}

\usepackage[utf8]{inputenc}
\usepackage{tabularx}

\usepackage{microtype}

\usepackage{inconsolata}

\usepackage{graphicx}

\usepackage{cleveref}

\usepackage{booktabs}
\usepackage{multirow}
\usepackage{xspace}

\usepackage{xcolor}
\usepackage{colortbl}
\colorlet{Gray}{gray!15}
\colorlet{DarkGray}{gray!33}
\newcommand{\cc}{\cellcolor{Gray}}
\usepackage[tight-spacing=true]{siunitx}
\def\:{\hskip0pt}

\usepackage{fvextra}
\usepackage{soul}
\colorlet{SoulYellow}{yellow!40!}
\sethlcolor{SoulYellow}

\newcommand{\titmn}{English-French Pun Translation Evaluation\xspace}
\renewcommand{\titmn}{Puns Lost in Translation\xspace}
\newcommand{\titsb}{Test Collection and Metric\xspace}
\renewcommand{\titsb}{A Location-Based Measure and Benchmark for English-French Pun Translation\xspace}

\title{\titmn: \titsb}

\author{Liana Ermakova \\
  University of Brest, HCTI\\
  Brest, France\\
  \texttt{Liana.Ermakova@ubo.fr} \\\And
  Igor Kuzmin \\ 
  Universitat Pompeu Fabra \& \\
  Barcelona Supercomputing Center \\
  Barcelona, Spain \\
  \texttt{email@domain} \\\And
  Jaap Kamps \\
  University of Amsterdam, ILLC \\
  Amsterdam, The Netherlands \\
  \texttt{kamps@uva.nl} \\}

\author{
  \textbf{Liana Ermakova}\textsuperscript{1\orcidlink{0000-0002-7598-7474}},
  \textbf{Igor Kuzmin}\textsuperscript{2\orcidlink{0009-0001-1513-1834}},
  \textbf{Jaap Kamps}\textsuperscript{3\orcidlink{0000-0002-6614-0087}}
\\
\\
  \textsuperscript{1}University of Brest, HCTI, Brest, France,\\
  \textsuperscript{2}Universitat Pompeu Fabra, Barcelona, Spain,\\
  \textsuperscript{3}University of Amsterdam, Amsterdam, The Netherlands\\
  \small{
    \textbf{Correspondence:} \href{mailto:liana.ermakova@univ-brest.fr}{liana.ermakova@univ-brest.fr}
  }
}

\begin{document}
\maketitle
\begin{abstract}
Existing machine translation measures and LLM-as-a-judge setups focus on global fluency and adequacy, so they fail to assess whether English–French translations preserve a pun or wordplay.
Pun translation is a hard testbed for creative generation, requiring reconstruction of ambiguity and wordplay, where literal but unfunny outputs can still score highly under current evaluation paradigms.
We introduce an annotated English-French pun test collection and a location-based measure to evaluate the preservation of the pun-bearing span, showing that it closely tracks human judgments and cleanly separates literal from successful translations, where standard measures and LLM judges do not.
Our results argue for location-aware, task-specific evaluation of creative MT and motivate extending location-based measures and resources to broader forms of wordplay, figurative language, and other language pairs.

\end{abstract}

\section{Introduction}
Recent advances in large language models (LLMs) have substantially improved
machine translation (MT), with systems achieving high fluency and semantic adequacy. However, these gains are uneven: pun translation, which requires preserving both meaning and wordplay, remains particularly challenging. Puns rely on ambiguity, phonetic resemblance, and culturally grounded associations between form and meaning. Their translation, thus, requires reconstructing a localized mechanism of wordplay that often depends on language-specific resources. As a result, literal translations tend to destroy the underlying pun, and in many cases the translation might become nonsensical (see \Cref{tab:translation_ex}). 

\begin{table*}[!t]
\small
\setlength\tabcolsep{0pt} 
\begin{tabular*}{\textwidth}{@{\extracolsep{\fill}} p{0.23\linewidth}>{\sl}p{0.24\linewidth}p{0.24\linewidth}>{\sl}p{0.24\linewidth}}
    \toprule
        \bf English pun & \bf Reference translation & \bf Location Explanation  & \bf Literal translation\\
        \midrule
        A skier retired because he was going downhill. & Le skieur est parti à la retraite. Il n'arrivait pas à remonter la pente. & “Remonter la pente” literally means to climb back up a slope and figuratively to recover from hardship. & Un skieur a pris sa retraite parce qu'il allait descendre.\\ 
        Scientists allow us to see the sun in a different light. & Les scientifiques nous permettent de voir le soleil sous une lumière différente. & “Voir sous une lumière différente” means to perceive something differently; literally, “in a different light.” & Les scientifiques nous permettent de voir le soleil d'une autre façon.\\
        Old bakers never die, they just keep making lots of dough. & Les vieux boulangers ne meurent jamais, ils continuent de se faire du blé. & “Se faire du blé” means to make money, "blé" literally means “wheat” & Les vieux boulangers ne meurent jamais, ils continuent de faire beaucoup de pâte.\\
    \bottomrule
\end{tabular*}
\caption{Examples of human and literal translations of English puns into French}
\label{tab:translation_ex}
\end{table*}

This makes pun translation require both specialized translation systems and evaluation methods. Current evaluation practices are poorly suited to this task: both standard measures and LLM-as-a-judge approaches focus on global sentence quality, allowing translations that fail to preserve wordplay to receive high scores.

To address this gap, we introduce both a dedicated evaluation resource and a measure for pun translation. We present a new English-French test set derived from the CLEF JOKER shared task \citep{DBLP:conf/clef/ErmakovaCBM25,DBLP:conf/clef/ErmakovaMBPSJ23,DBLP:conf/sigir/ErmakovaBJM23}, combining human translations and system outputs from multiple campaigns. The data are manually annotated by native French, trained translators for (1) meaning preservation and (2) wordplay presence, enabling fine-grained evaluation.

\begin{figure}[!t]
     \centering
     \includegraphics[width=\linewidth]{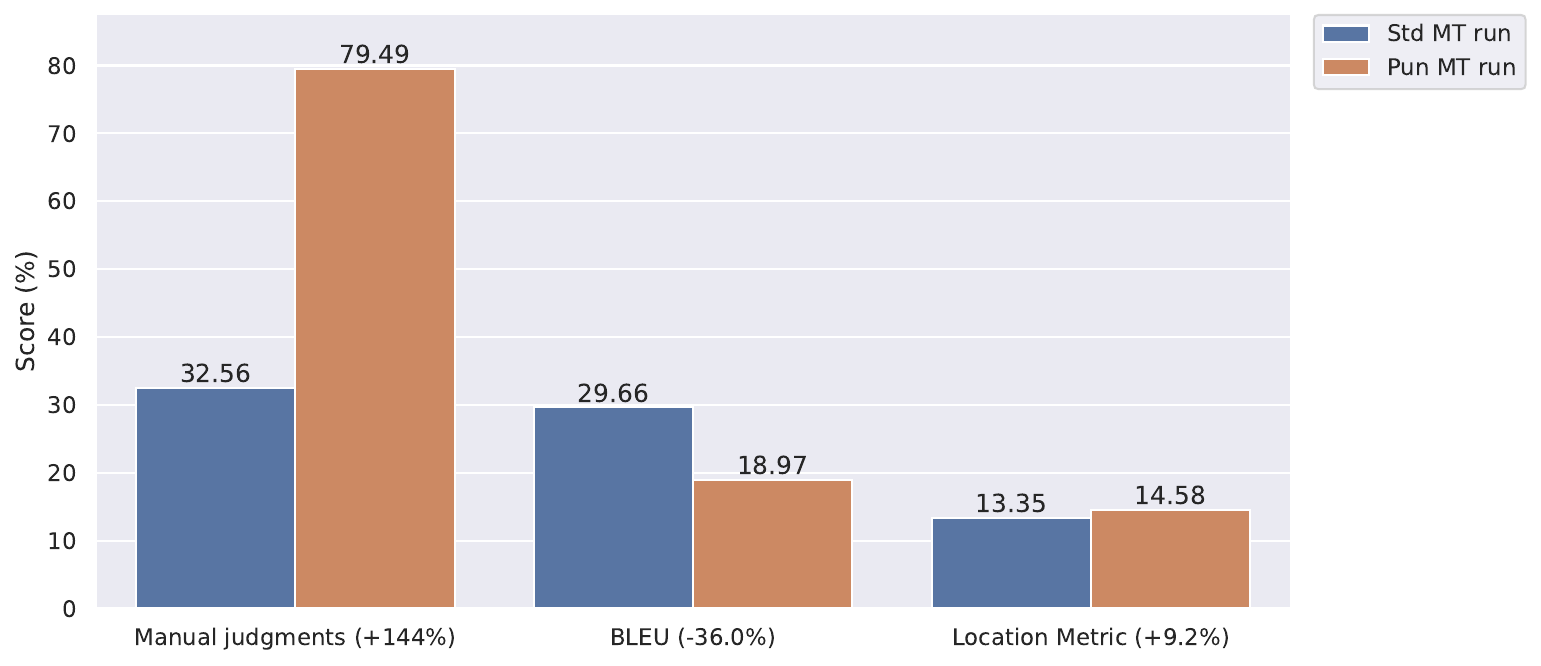}
     \caption{Standard MT and humor-aware pun translation evaluation: manual judgments show more literal translations are no wordplay, yet even against wordplay-preserving references, BLEU prefers literal translations.}
     \label{fig:teaser}
\end{figure}
Remarkably, even with a set of human wordplay-preserving references, standard evaluation measures tend to prefer literal translations that fail to preserve wordplay. \Cref{fig:teaser} illustrates this deviation between the BLEU score against reference translations and direct manual judgments.
This results from the greater variation required to preserve wordplay, and that the literal parts of the translation are identical across multiple human reference translations. Hence, the main difference is in the correct translation of the pun word. 

Based on this finding, we propose a location-based measure that evaluates whether the pun-bearing span is preserved, complementing global measures, and evaluate it on two subcorpora against state-of-the-art supervised and unsupervised MT measures. \Cref{fig:teaser} shows this measure against the set of human reference translations, scoring the pun-aware run 9.2\% higher than the literal translation run.
In addition, we evaluate LLMs-as-a-judge for pun translation using two prompting strategies: holistic quality judgments and a decomposition into meaning preservation and wordplay presence.

Our contributions are threefold: (i) a large-scale, manually annotated test collection for English-French pun translation used in the CLEF JOKER track \cite{DBLP:conf/clef/ErmakovaKVMBK26,DBLP:conf/clef/ErmakovaNBDK26}, (ii) a new evaluation measure based on pun location, and (iii) an empirical analysis of LLM-as-a-judge performance on the pun translation task. 

The paper is organized as follows. \Cref{sec:sota} presents existing MT evaluation.
\Cref{sec:method} describes the corpus, the location-based measure, and two LLM-as-a-judge strategies. \Cref{sec:eval} describes the subcorpora used for evaluation, along with the baseline MT measures. \Cref{sec:res} analyzes LLM-as-a-judge results and compares our measure to baselines. \Cref{sec:conclusion} discusses our findings.

\section{Related Work}
\label{sec:sota}

Following the MT-evaluation taxonomy of \citet{DBLP:journals/csur/MukherjeeS26}, we distinguish three survey-level families (unsupervised, supervised, and LLM-based) and add a task-specific family for pun-bearing span preservation (see \Cref{tab:metric-taxonomy}).

\begin{table*}[!t]
\small
\setlength\tabcolsep{0pt} 
\begin{tabular*}{\textwidth}{@{\extracolsep{\fill}} p{0.13\linewidth}p{0.14\linewidth}p{0.18\linewidth}p{0.12\linewidth}p{0.14\linewidth}p{0.22\linewidth}}
\toprule
\textbf{Family} & \textbf{Modeling View} & \textbf{Learning Signal} & \textbf{Reference Use} & \textbf{Typical Output} & \textbf{Role} \\
\midrule
Unsupervised
& Lexical, character, and embedding similarity
& No measure\:-\:specific supervised training
& Usually reference\:-\:based
& Similarity or distance score
& Family\:-\:level baseline for adequacy/fluency\:-\:oriented signals. \\
\cmidrule(lr){1-6}
Supervised
& Learned quality estimation/regression
& Human ratings and/or error annotations
& Reference\:-\:based or reference\:-\:free
& Predicted quality score (optionally spans)
& Strong learned baseline family across reference settings. \\
\cmidrule(lr){1-6}
LLM\:-\:based
& Prompted judge with rubric\:-\:conditioned reasoning
& Instruction\:-\:following judgment
& Optional reference
& Scalar score with optional rationale
& Flexible rubric\:-\:sensitive family evaluated for agreement behavior. \\
\cmidrule(lr){1-6}
Task\:-\:specific (Location\:-\:based Measure)
& Pun\:-\:location structure preservation
& Pun\:-\:location annotations from JOKER and extended references
& Source\:-\:target alignment required
& Location\:-\:based preservation score
& Hypothesized complementary signal for pun transfer; tested empirically later. \\
\bottomrule
\end{tabular*}
\caption{Family\:-\:level taxonomy, including unsupervised, supervised, LLM\:-\:based, and task\:-\:specific (location\:-\:based) families. The table defines conceptual families and study roles, not empirical findings or rankings.}
\label{tab:metric-taxonomy}
\end{table*}

\subsection{Unsupervised measures}

Unsupervised measures evaluate translations without task-specific fitting to human quality judgments. In standard MT-evaluation taxonomies, this family includes lexical overlap between candidate and reference translations, character-level similarity, and embedding-based similarity measures. BLEU \citep{papineni-etal-2002-bleu} measures word n-gram overlap with a reference, chrF \citep{popovic-2015-chrf} emphasizes character-level precision and recall and is often more robust to inflectional variation, and TER measures
the normalized minimum number of edits needed to transform a machine translation into a reference translation \citep{snover-etal-2006-study}. Beyond surface overlap, BERTScore compares contextual token representations and therefore captures semantic similarity more flexibly than exact matching \citep{DBLP:conf/iclr/ZhangKWWA20}. These measures are attractive because they are efficient, widely adopted, and relatively easy to interpret. At the same time, they assess global translation quality, especially semantic overlap between source and target, rather than preserving localized wordplay ambiguity.
This limitation is consistent with prior work in computational humor, which argues that surface-level similarity is often a poor proxy for humor preservation and humorous effect \citep{su2025psychology}.

\subsection{Supervised measures}

Supervised measures have become the state-of-the-art for correlating with human judgment. Our evaluation includes neural reference-based measures BLEURT-20 \citep{sellam-etal-2020-bleurt}, xCOMET-XL \citep{guerreiro-etal-2024-xcomet}, COMETKiwi-XL \citep{rei-etal-2023-scaling}, and BLASER \citep{dale-costa-jussa-2024-blaser}. 
BLEURT-20~\citep{pu-etal-2021-learning} is a RemBERT-based regressor fine-tuned on WMT human ratings that scores a hypothesis against a reference. 
xCOMET-XL is a reference-based COMET model that additionally predicts error spans. 
COMETKiwi-XL is a reference-free (quality-estimation) COMET model that scores only the source-hypothesis pair. 
BLASER~2.0 compares SONAR sentence embeddings of the source, the hypothesis and, optionally, the reference, in both reference-based and reference-free variants. 

While these models are robust to semantic variation, they are typically trained on general MT error corpora rather than creative or figurative text. Specialized supervised measures like MMTE \citep{wang2024mmte} have been proposed for metaphorical language, not humor specifically, but standard supervised models remain the primary benchmarks for cross-domain quality estimation.

Overall, supervised measures usually correlate better with human judgments of translation quality than unsupervised measures. However, they are still optimized for general MT quality rather than explicit pun preservation. A translation can receive a high score for fluency and adequacy while losing the original pun's locus or mechanism.

\subsection{LLM-as-a-judge}
\label{sec:sota:llmaajudge}

The LLM-as-a-judge paradigm is increasingly used across NLP as a proxy for human evaluation \citep{gu_survey_2026,balog_sigir_2025,llm_rel_sigir2023}. It offers flexible, rubric-conditioned scoring, but prior work reports consistent limitations: LLM judges can be more lenient than humans, less sensitive to small but meaningful system differences, biased toward longer outputs, and sometimes favorable to outputs from the same model family \citep{balog_sigir_2025,dubois2024lengthAlpacaEval,liu-etal-2024-llms-narcissistic}. 

Recent humor-oriented frameworks such as GEMBA-MQM provide global humor scoring, yet they still offer limited observability of where the pun mechanism is preserved or lost \citep{su2025psychology}. More broadly, recent work on humor understanding also shows that even strong LLMs struggle with reliably identifying humorous punchlines, underscoring the difficulty of using global model judgments as a proxy for fine-grained humorous effect \citep{romanowski2025punchlines}. Evidence about broader bias toward AI-generated text remains mixed \citep{balog_sigir_2025}. Prior work has shown that LLM-as-a-judge approaches can exhibit sycophantic behavior, reinforcing premises embedded in the query, particularly for non-factoid tasks such as humor interpretation \citep{DBLP:conf/chiir/ErmakovaFK26}. 
Recent work has probed the robustness of MT evaluation
using ambiguity-sensitive perturbations such as Chinese homophones, suggesting that evaluation models may be vulnerable to subtle meaning-preservation failures in user-generated text \citep{qian2025homophones}. 

In this paper, we assess LLM-as-a-judge for pun translation using two prompting strategies -- holistic quality judgments and a decomposition into meaning preservation and wordplay presence -- to examine whether LLM-based evaluation can distinguish between fluent but literal translations and those that successfully preserve wordplay.

\subsection{Why a location-based measure is needed}
Across unsupervised, supervised, and LLM-based families, most scores primarily capture global adequacy and fluency.
However, pun translation depends on preserving a localized structural pivot, and that mechanism can be lost even when global scores remain high. This gap is not specific to surface measures: even structured error-based LLM judges such as GEMBA-MQM rely on an MQM taxonomy with no category for wordplay loss, so a literal translation that destroys the pun incurs no penalty. A location-based measure addresses this
gap by explicitly scoring whether the annotated pun-bearing span is preserved in translation.

\section{Method}
\label{sec:method}

This section presents the corpus, the location-based measure, and two LLM-as-a-judge strategies. 

\subsection{Corpus}
The corpus is built based on data from the Pun translation task of the CLEF JOKER track~\citep{DBLP:conf/clef/ErmakovaCBM25,DBLP:conf/clef/ErmakovaMBPSJ23,DBLP:conf/sigir/ErmakovaBJM23}. The JOKER corpus is a French translation of the English corpus for pun detection and interpretation  \cite{miller-etal-2017-semeval}. 

The dataset includes French translations produced by professional translators, native French speakers, and automatic runs submitted to the JOKER track in 2023 and 2025. Native French translators manually annotated machine translations for meaning preservation and wordplay, covering 13,755 distinct French translations of 2,480 English puns from the 2023 runs and 3,751 French translations of 1,665 English puns from the 2025 runs. All 12,861 French translations containing wordplay from 4,059 distinct English puns are manually annotated with the words or phrases carrying multiple meanings (7,786 distinct pun locations). Among them, 8,823 were produced by professional translators (native French speakers), and 4,038 were successful automatic translations (i.e., preserving meaning and containing wordplay). We report the results only on a subset of the data detailed in \Cref{sec:eval}.

We measured agreement after adding location spans to 2,087 wordplay instances: only 38 were reclassified as non-wordplay, corresponding to 98.18\% agreement. Besides, two native French-speaking master's students in translation independently annotated 598 French translations of 93 English puns. We used Gwet's AC1 \cite{gwet_computing_2008} to avoid the prevalence paradox, as Cohen's kappa can be low despite high observed agreement when one category predominates. Since most translations were literal, the predominant label was “no wordplay” for wordplay and “meaning preserved” for meaning preservation. Gwet's AC1 reached 0.838 for meaning preservation and 0.898 for wordplay classification (see \Cref{appendix:agreement} for details).

\subsection{Location-based measure}
Pun location-based evaluation allows for a more fine-grained analysis of MT. We identified words or phrases with multiple meanings (pun locations) in the reference texts by combining French reference translations with pun location annotations. We report the proportion of candidate translations that contain at least one annotated pun-location token among the corresponding reference sentences. 

\Cref{tab:location_ex} shows example reference translations for an English pun along with their location annotations. A candidate translation is considered correct if it contains one of the annotated locations that realizes a double meaning with the photography terminology \emph{\{'flous', 'mise au point', 'mises au point', 'trépieassent', 'pellicules', 'pas pareil', 'objectif', 'objectifs', 'retouchent'\}}. 

\begin{table}[!t]
\footnotesize
\setlength\tabcolsep{0pt} 
\begin{tabular*}{\linewidth}{@{\extracolsep{\fill}} p{\linewidth}}
   \toprule
        \bf English pun / References + \hl{Location}\\
        \midrule
OLD PHOTOGRAPHERS never die, they just lose their \hl{focus}.\\
\cmidrule(lr){1-1}
\sl Un photographe ne meurt jamais, seulement, il n'a plus d'\hl{objectifs}.\\\sl Un photographe, ça ne meurt pas, ça change seulement d'\hl{objectif}.\\\sl Quand les vieux photographes meurent, ils \hl{trépieassent}.\\\sl Les vieux photographes ne meurent jamais, ils \hl{retouchent} du bois pour rester en vie.\\\sl Les vieux photographes ne meurent pas, ils perdent leurs \hl{objectifs} de vue.\\\sl Les photographes ne meurent jamais, ils vont juste faire des \hl{mises au point}.\\\sl Les vieux photographes ne meurent pas, mais ils attrapent des \hl{pellicules}.\\\sl Quand un photographe vieillit, il ne meurt pas, mais quand même, c’est \hl{pas pareil}.\\\sl Les vieux photographes perdent souvent vue de leurs objectifs et ont besoin d'une \hl{mise au point}.\\\sl Les photographes ne meurent pas, seulement, ils deviennent des vieux \hl{flous}.\\    \bottomrule
    \end{tabular*}
\caption{Examples of location annotation}
\label{tab:location_ex}
\end{table}

\subsection{LLM-as-a-judge}
To evaluate LLM-as-a-judge on pun translation, we conducted two prompting experiments with the state-of-the-art models \emph{GPT-5.1}, \emph{Gemini-3-flash-preview}, \emph{Qwen3-4B-Instruct}, and \emph{Meta-Llama-3.1-8B-Instruct} (see models' parameters in \Cref{appendix:llm_param}). 
In the first experiment, we directly asked the models to evaluate translation quality (exact models' prompt in \Cref{appendix:llm_prompts}). 

In the second experiment, we mimic our human evaluation by independently assessing two aspects of the pun translation: meaning preservation and the presence of wordplay in the French version. A translation is considered successful if it preserves the meaning and contains wordplay (exact models' prompts in \Cref{appendix:llm_prompts}). 

We post-processed LLM outputs to homogenize them (see \Cref{appendix:postprocessing}) and then compared the predictions with human annotations.

\section{Evaluation Framework}
\label{sec:eval}

This section describes the two subcorpora used for evaluation, along with the baseline MT measures. 

\subsection{Subcorpora}
We conducted the experiments on two subcorpora: (1) the JOKER 2025 subcorpus, consisting of participant-submitted translations manually annotated for meaning preservation and the presence of wordplay, combined with the set of corresponding manually created references (hereafter 2025); and (2) a mixed subcorpus constructed as described below (hereafter 2023). The 2025 subcorpus is evaluated against
reference translations and compared to the manual annotation of a sample of 50 predictions for each submission.  

The 2023 subcorpus is constructed to have human annotations for detailed analysis.  
Specifically, we intersected runs via their corresponding English source puns. The resulting dataset includes: (1) the subset of runs submitted by JOKER participants in 2023 that were manually annotated for meaning preservation and the presence of wordplay; (2) the best-scoring run from 2025 (hereafter DSGT), which we manually annotated; and (3) an oracle run based on translations produced by professional French translators. 

The DSGT run is based on multi-agent translation with contrastive learning and phonetic-semantic embeddings, and trained on the dataset provided by the JOKER organizers~\citep{dsgt}. DSGT translations were often non-literal and judged successful in about 77\% of cases (after full annotation on all 2023 data included also in the 2025 test batch). In contrast, the runs submitted in 2023 produced only literal translations, of which approximately 17\% were judged successful by the annotators out of 13,755 evaluated translation pairs~\cite{DBLP:conf/clef/ErmakovaMBPSJ23c}. 

To simulate systems with varying performance, we generated synthetic runs from the oracle run by replacing a controlled proportion of its translations with literal translations from the 2023 runs (10–90\%). Note that as we intersected submitted runs with the references, the proportion of the successful translations is higher than the average 17\% in 2023~\cite{DBLP:conf/clef/ErmakovaMBPSJ23c}. Only manually created translations were retained as references; successful automatic translations were discarded. The manual translations used to build the oracle run were excluded from the reference set. Consequently, the second subcorpus contains only high-quality manual reference translations, annotated by a different native French translator. 

\Cref{tab:subcorpora_stats} reports the statistics of both datasets, including the number of distinct English source puns (\#EN), distinct French translations (\#FR), references (\#refs), distinct locations (\#loc), and the number of manually annotated pairs (\#man).

\begin{table}[!t]
    \small
    \begin{tabular*}{\linewidth}{@{\extracolsep{\fill}} lrrrrr}
    \toprule
         Corpus & \#EN & \#FR &\#refs & \#loc & \#man\\
    \midrule
         2025 & 2,277 & 15,164  & 2,466 & 2,309 & 1,120\\
         2023 & 307 & 919 & 1,123 & 715 & 919  \\ 
    \bottomrule
    \end{tabular*}
    \caption{Statistics of the 2023 and 2025 subcorpora}
    \label{tab:subcorpora_stats}
\end{table}
\subsection{Baselines}
\label{sec:baselines}

We evaluate four baseline families. The unsupervised baselines are BLEU, chrF++, TER, and BERTScore. BLEU measures $n$-gram overlap with a reference~\citep{papineni-etal-2002-bleu}; chrF++ measures character-level precision and recall and is more tolerant to inflectional variation~\citep{popovic-2015-chrf}; TER measures edit distance as a proxy for post-editing effort~\citep{snover-etal-2006-study}; and BERTScore compares contextual token embeddings rather than exact surface overlap~\citep{DBLP:conf/iclr/ZhangKWWA20}. For BLEU, chrF++, and TER we use the sacreBLEU implementation~\citep{post-2018-call}; for BERTScore we use the standard \texttt{bert-score} implementation.

Our supervised baselines cover both reference-based and quality-estimation settings: BLEURT-20~\citep{sellam-etal-2020-bleurt}, XCOMET-XL~\citep{guerreiro-etal-2024-xcomet}, COMETKiwi-XL~\citep{rei-etal-2023-scaling}, and BLASER 2.0 in both reference-based and QE variants~\citep{dale-costa-jussa-2024-blaser}. BLEURT-20, XCOMET-XL, and BLASER Ref use references at scoring time, whereas COMETKiwi-XL and BLASER QE score a source--hypothesis pair without
references. All supervised measures use publicly released checkpoints without any fine-tuning on JOKER data.

\begin{table*}[!t]
    \small
    \setlength\tabcolsep{0pt} 
    \begin{tabular*}{\linewidth}{@{\extracolsep{\fill}} lccccccccccccccccccc}
    \toprule
    \multirow{3}{*}{Prompt} & \multicolumn{14}{c}{2023} & \multicolumn{5}{c}{2025}\\
    \cmidrule(lr){2-15} \cmidrule(lr){16-20}
    &  \multicolumn{2}{c}{\cc Manual} & \multicolumn{3}{c}{Gemini} & \multicolumn{3}{c}{GPT} & \multicolumn{3}{c}{Qwen}& \multicolumn{3}{c}{Llama}& \cc Manual & Gemini & GPT & Qwen & Llama\\
    \cmidrule(lr){2-3} \cmidrule(lr){4-6}\cmidrule(lr){7-9} \cmidrule(lr){10-12} \cmidrule(lr){13-15}\cmidrule(lr){14-14}\cmidrule(lr){15-15}\cmidrule(lr){16-16}\cmidrule(lr){17-17}\cmidrule(lr){18-18}\cmidrule(lr){19-19} \cmidrule(lr){20-20}
    & \cc auto & \cc dsgt & ref & auto & dsgt& ref & auto & dsgt& ref & auto & dsgt& ref & auto & dsgt & \cc runs & runs & runs& runs& runs\\
         \midrule
        Meaning & \cc 67 & \cc 91 & 37 & 55 & 63 &
        38 & 51 & 54 & 31 & 49 & 29 & 0 & 0 & 8 & \cc 36 &
        46 & 52 & 42 & 13\\
        Wordplay & \cc 37 & \cc 79 & 96 & 50 & 78 &
        87 & 54 & 77 & 97 & 41 & 78 & 88 & 43& 77 & \cc 44 & 55 &
        70 & 49 & 57\\
        \midrule
        Good & \cc 36 & \cc 77 & 42 & 70 & 63 &
        35 & 68 & 60 & 43 & 60  & 52 & 3 & 64 & 24 & \cc 28 & 70 &
        69 & 63 & 72\\
        \bottomrule
    \end{tabular*}
     \caption{LLM-as-a-judge accuracy (\%): agreement of LLM-as-a-judge with manual labels (shown in Manual)}
    \label{tab:llm_as_judge_results}
\end{table*}

As LLM baselines, we use \emph{GPT-5.1}, \emph{Gemini-3-flash-preview}, \emph{Qwen3-4B-Instruct}, and \emph{Meta-Llama-3.1-8B-Instruct} with the two prompting strategies described in the LLM-as-a-judge subsection above, and additionally GEMBA-MQM~\citep{kocmi-federmann-2023-gemba-mqm}, a structured error-span LLM judge, applied with
\emph{GPT-5.1} and \emph{Gemini-3-flash-preview} using the providers' default sampling settings (GPT-5.1, as a reasoning model, does not accept a temperature parameter; for Gemini-3 we keep the recommended default temperature of 1.0 with a low thinking level).

We select GEMBA-MQM as a representative state-of-the-art LLM judge because it operates at the sentence level, requires no reference, and is directly comparable to our holistic and decomposed prompts.\footnote{We do not run its multi-run successor GEMBA-MQM V2~\citep{junczys-dowmunt-2025-gemba-v2}, M-Prometheus~\citep{pombal2025mprometheus}, TREQA~\citep{fernandes2025treqa}, or LiTransProQA~\citep{zhang-etal-2025-litransproqa} as baselines. V2 differs from GEMBA-MQM only in stochastic multi-run aggregation while sharing the same MQM error taxonomy, which contains no category for wordplay loss, and has the same limitation we probe here. M-Prometheus, TREQA, and LiTransProQA discussed in Section~\ref{sec:sota:llmaajudge} are designed for paragraph-level literary text, and their question-answering or long-form judgment procedures do not transfer to single-sentence
context-free puns.}
Our LLM-based baselines instantiated with frontier commercial and open-weight models thus cover the three main sentence-level LLM-judge paradigms---holistic direct assessment, decomposed rubric-based judgment, and structured error-span annotation (GEMBA-MQM).

Aggregation is done at the sentence level first and at the run
level second. For each system output, we align the translation with
all references available for the same English source sentence. BLEU,
chrF++, and TER are computed on \emph{corpus-level}  in a
multi-reference setting using sacreBLEU~\citep{post-2018-call}. 
Since the number of references varies per source pun, sources with fewer references than the maximum are scored against their available references only.  TER is reported on sacreBLEU's standard percentage scale (edit operations per reference word $\times$ 100), where scores above 100 are possible for hypotheses requiring more edits than the reference length. For pairwise reference-based measures (BLEURT-20, XCOMET-XL, and BLASER~Ref) we score the candidate against each available reference and retain the maximum score for that source sentence. Run-level scores are then the \emph{mean} of these per-sentence maxima. BERTScore (F1) is computed against the first available reference of each source
and averaged at the run level.\footnote{On the 2023 subcorpus, whose oracle run consists of professional translations, we remove the oracle translations (up to punctuation and whitespace) from the reference set of each source, so that no run is scored against its own output.}  
Reference-free measures (COMETKiwi-XL, BLASER~QE, and GEMBA-MQM) are computed once per source-translation pair and
averaged at the run level. 
For GEMBA-MQM, each segment receives a penalty based on the errors the LLM judge identifies -- critical ($w{=}25$), major ($w{=}5$), minor ($w{=}1$) -- truncated to the first five errors following the original protocol, yielding a raw score $s_{\mathrm{raw}}\in[-25,0]$, which we linearly rescale to $[0,100]$ via $s_{\mathrm{norm}} = 100 + 4\,s_{\mathrm{raw}}$
(100 = no errors), and we report $s_{\mathrm{norm}}$ as the main score.

\section{Results}
\label{sec:res}

This section analyzes LLM-as-a-judge results and compares our measure to baselines.

\subsection{LLM-as-a-judge}

\Cref{tab:llm_as_judge_results} reports the accuracy (in percentage) of the LLM-as-a-judge predictions—meaning preservation, wordplay presence, and good translation—on the 2023 and 2025 subcorpora, compared against human annotations. \Cref{tab:llm_as_judge_results} also shows the percentage of translations in the runs that were judged successful along these dimensions based on manual evaluation (columns \emph{Manual}). For the 2023 subcorpus, we report accuracy for the manual references, the automatic 2023 run, and the best-scoring 2025 run DSGT ~\cite{dsgt}. Human annotations include only meaning preservation and wordplay presence. The good translation prediction is evaluated against successful translations, defined as translations that both preserve meaning and contain wordplay. 

\begin{table}[!t]
\small
\setlength{\tabcolsep}{0pt}
    \begin{tabular*}{\linewidth}{
        @{\extracolsep{\fill}} lr
        S[table-format=3.0,round-mode=places,round-precision=0]
        S[table-format=3.0,round-mode=places,round-precision=0]
        S[table-format=3.0,round-mode=places,round-precision=0]
        S[table-format=3.0,round-mode=places,round-precision=0]
    }
         \toprule
         run & Manual & {Gemini} & {GPT} & {Qwen} & {Llama}\\
         \midrule
         Oracle (ref) & 100 &  95.76547231270358 &  87.29641693811075 &  96.74267100977198 &  87.94788273615634\\
         auto & 37 & 85.99348534201955 &  74.2671009771987 &  92.83387622149837 &  81.10749185667753\\
         dsgt & 79 & 99.67105263157895 &  97.36842105263158 &  99.3421052631579 &  97.69736842105263\\
         \bottomrule
    \end{tabular*}
    \caption{\% of translations judged as wordplay (2023 subcorpus)}
    \label{tab:wordplay_yes_2023}
\end{table}

\begin{table}[!t]
\small
\setlength{\tabcolsep}{0pt}
       \begin{tabular*}{\linewidth}{
        @{\extracolsep{\fill}} l
        S[table-format=3.0,round-mode=places,round-precision=0]
        S[table-format=3.0,round-mode=places,round-precision=0]
        S[table-format=3.0,round-mode=places,round-precision=0]
        S[table-format=3.0,round-mode=places,round-precision=0]
        S[table-format=3.0,round-mode=places,round-precision=0]
        S[table-format=3.0,round-mode=places,round-precision=0]
        S[table-format=3.0,round-mode=places,round-precision=0]
        S[table-format=3.0,round-mode=places,round-precision=0]
        S[table-format=3.0,round-mode=places,round-precision=0]
    }
        \toprule
        {Refs} & {Man} & \multicolumn{2}{c}{Gemini} & \multicolumn{2}{c}{GPT}& \multicolumn{2}{c}{Qwen} & \multicolumn{2}{c}{Llama}\\
        \cmidrule(lr){3-4} \cmidrule(lr){5-6}\cmidrule(lr){7-8} \cmidrule(lr){9-10}
        & & {No} & {Yes} & {No} & {Yes} & {No} & {Yes} & {No} & {Yes} \\
        \midrule
No & 55.625 & 14.107142857142858 & 41.517857142857146 & 30.80357142857143 & 24.821428571428573 & 5.178571428571429 & 50.44642857142857 & 15.0 & 40.625\\
Yes & 44.375 & 3.303571428571429 & 41.07142857142857 & 5.178571428571429 & 39.19642857142857 & 0.7142857142857143 & 43.660714285714285 & 2.3214285714285716 & 42.05357142857143\\
All & 100.0 & 17.410714285714285 & 82.58928571428571 & 35.982142857142854 & 64.01785714285714 & 5.892857142857142 & 94.10714285714286 & 17.32142857142857 & 82.67857142857142\\
\bottomrule
    \end{tabular*}
    \caption{LLM confusion matrices (\%) for French wordplay detection (2025 subcorpus)}
    \label{tab:llm_crosstable}
\end{table}

On the 2023 subcorpus, LLMs detect wordplay in the reference translations with very high accuracy: Gemini 96\%, GPT 87\%, Qwen 97\%, and Llama 88\%. However, performance decreases on system outputs, with both models showing lower accuracy on the DSGT run and the automatic literal run. \Cref{tab:wordplay_yes_2023} provides further insight by showing that LLMs tend to label wordplay much more frequently than humans. On the 2025 subcorpus, the accuracy of LLM wordplay detection in French is between 49\% and 70\%. Human annotators identify wordplay in 44\% of translations, compared to 83\% for Gemini, 64\% for GPT, 94\% for Qwen, and 83\% for Llama (see \Cref{tab:llm_crosstable}). This behavior may reflect the extreme sycophancy reported in previous work on wordplay detection~\cite{DBLP:conf/chiir/ErmakovaFK26}. Examples of the effect of prompt framing on LLM judgments about wordplay are given in \Cref{app:sycophancy_judge}. Alternatively, the models may rely on the presence of a source English pun and therefore assume that its translation also contains wordplay. 

\begin{table}[!t]
    \small
    \setlength\tabcolsep{0pt} 
    \begin{tabular*}{\linewidth}{
        @{\extracolsep{\fill}} l
        S[table-format=3.0,round-mode=places,round-precision=0]
        S[table-format=3.0,round-mode=places,round-precision=0]
        S[table-format=3.0,round-mode=places,round-precision=0]
        S[table-format=3.0,round-mode=places,round-precision=0]
        S[table-format=3.0,round-mode=places,round-precision=0]
        S[table-format=3.0,round-mode=places,round-precision=0]
    }
    \toprule
\multirow{2}{*}{Model} & \multicolumn{3}{c}{Oracle (ref)} & \multicolumn{3}{c}{Literal}\\
\cmidrule(lr){2-4} \cmidrule(lr){5-7}
& {No} & {Partially} & {Yes}& {No} & {Partially} & {Yes} \\
\midrule
Gemini & 19.54397394136808 & 42.99674267100978 & 37.45928338762215 & 74.5928338762215 &  & 25.407166123778502  \\
GPT & 42.01954397394137 & 20.195439739413683 & 37.785016286644954 & 53.74592833876222 & 18.892508143322477 & 27.36156351791531  \\
Qwen & 41.042345276872965 & 27.68729641693811 & 31.27035830618892 & 37.13355048859935 & 29.64169381107492 & 33.22475570032573 \\
Llama &  & 100.0 &  &  & 99.3485342019544 & \\
\midrule
Manual & & & 100 & 33 & & 67\\
\bottomrule
\end{tabular*}
     \caption{LLM predictions of meaning preservation (\%) on Oracle and literal translations (2023 subcorpus)}
    \label{tab:llm_as_judge_meaning}
\end{table}

\begin{table}[!t]
    \small
    \setlength\tabcolsep{0pt} 
    \begin{tabular*}{\linewidth}{
        @{\extracolsep{\fill}} l
        S[table-format=3.0,round-mode=places,round-precision=0]
        S[table-format=3.0,round-mode=places,round-precision=0]
        S[table-format=3.0,round-mode=places,round-precision=0]
        S[table-format=3.0,round-mode=places,round-precision=0]
    }
    \toprule
\multirow{2}{*}{Model} & \multicolumn{2}{c}{Oracle (ref)} & \multicolumn{2}{c}{Literal}\\
\cmidrule(lr){2-3} \cmidrule(lr){4-5}
&{No} & {Yes}& {No} & {Yes}\\
\midrule
Gemini & 57.65472312703584 & 42.34527687296417 & 72.9641693811075 & 27.035830618892508\\
GPT & 64.82084690553745 & 35.17915309446254 & 77.52442996742671 & 22.475570032573287\\
Qwen & 57.32899022801303 & 42.671009771986974 & 56.02605863192183 & 43.97394136807817\\
Llama & 97.06840390879479 & 2.9315960912052117 & 97.06840390879479 & 2.6058631921824107\\
\midrule
Manual &  & 100 & 64 & 36\\
\bottomrule
\end{tabular*}
     \caption{LLM predictions of good translations (\%) for Oracle and literal translations (2023 subcorpus)}
    \label{tab:llm_as_judge_good}
\end{table}

\begin{table*}[!t]
\small
\setlength\tabcolsep{1.5pt}
         \begin{tabular*}{\textwidth}{@{\extracolsep{\fill}} l*{1}{S[table-format=3.1,round-mode=places,round-precision=1]}r*
         {8}{S[table-format=2.2,round-mode=places,round-precision=2]}
         S[table-format=2.2,round-mode=places,round-precision=2]
S[table-format=2.2,round-mode=places,round-precision=2]
S[table-format=2.1,round-mode=places,round-precision=1]
S[table-format=2.1,round-mode=places,round-precision=1]}
         \toprule
         {\multirow{2}{*}{run\_id}} & {\multirow{2}{*}{\%man.\(\uparrow\)}} & {\multirow{2}{*}{\#Loc.\(\uparrow\)}} & {\multirow{2}{*}{\%Loc.\(\uparrow\)}} & {\multirow{2}{*}{BLEU\(\uparrow\)}} & {\multirow{2}{*}{chrF\(\uparrow\)}} & {\multirow{2}{*}{TER\(\downarrow\)}} & {\multirow{2}{*}{BERT\(\uparrow\)}} & {\multirow{2}{*}{BLEURT\(\uparrow\)}} & \multicolumn{2}{c}{XCOMET\(\uparrow\)} & \multicolumn{2}{c}{BLASER\(\uparrow\)} & \multicolumn{2}{c}{GEMBA-MQM\(\uparrow\)}\\
         \cmidrule(lr){10-11}\cmidrule(lr){12-13}\cmidrule(lr){14-15}
         &&&&&&&&&{XL} & {KIWI} & {REF} & {QE} & {GPT-5.1} & {Gemini-3}\\
         \midrule
{Oracle} & 100 & 184 & 59.93485342 & 40.3691 & 57.1376 & 55.1658 & 84.6620 & 58.0627 & 76.8519 & 50.40201779 & 4.0911 & 4.178873345 & 57.40716612 & 75.62214984\\
{10\%} & 91.20521172638436 & 180 & 58.63192182 & 40.2732 & 57.0147 & 55.6270 & 84.5772 & 57.2407 & 76.5319 & 50.17715286 & 4.0866 & 4.18344841 & 56.84690554 & 74.44951140\\
{20\%} & 86.97068403908796 & 177 & 57.65472313 & 39.8940 & 56.3780 & 56.5238 & 84.4798 & 57.0074 & 76.4193 & 50.06983341 & 4.0994 & 4.191362709 & 55.97394137 & 74.87947883\\
{30\%} & 77.85016286644951 & 168 & 54.72312704 & 39.1549 & 56.5111 & 55.7808 & 84.5525 & 56.1861 & 76.6396 & 49.57200825 & 4.1156 & 4.219634872 & 52.95114007 & 73.92833876\\
{40\%} & 71.66123778501628 & 158 & 51.46579805 & 38.6683 & 55.9715 & 56.8313 & 84.5269 & 56.3545 & 76.5645 & 50.20028616 & 4.1150 & 4.212518464 & 52.62540717 & 73.26384365\\
{50\%} & 67.42671009771986 & 159 & 51.79153094 & 39.3603 & 56.0712 & 56.6007 & 84.5049 & 55.9927 & 76.6545 & 49.9279754 & 4.1447 & 4.25105132 & 54.51465798 & 73.99348534\\
{60\%} & 61.563517915309454 & 146 & 47.55700326 & 39.5616 & 56.3505 & 56.2164 & 84.2994 & 55.9094 & 76.5838 & 50.70033228 & 4.1694 & 4.269571691 & 53.86319218 & 72.36482085\\
{70\%} & 55.37459283387622 & 146 & 47.55700326 & 38.3063 & 55.0405 & 56.7544 & 84.2430 & 54.7976 & 76.2877 & 49.30800245 & 4.1188 & 4.236517406 & 52.82084691 & 71.14006515\\
{80\%} & 48.85993485342019 & 149 & 48.53420195 & 39.8974 & 56.2872 & 55.2683 & 84.5120 & 55.7098 & 77.1148 & 51.75506826 & 4.1878 & 4.301404379 & 54.26710098 & 73.44625407\\
{90\%} & 41.69381107491857 & 140 & 45.60260586 & 37.8511 & 54.3191 & 57.2669 & 84.1299 & 54.6806 & 76.6731 & 50.41549099 & 4.1684 & 4.290257686 & 53.64169381 & 72.19543974\\
{Auto} & 35.83061889250814 & 135 & 43.97394137 & 37.9563 & 54.5452 & 57.4975 & 84.1051 & 54.0437 & 76.4528 & 50.30099116 & 4.1778 & 4.30238266 & 52.65146580 & 71.03583062\\
{DSGT} & 77.19869706840392 & 156 & 50.81433225 & 27.7270 & 48.5370 & 68.1309 & 81.0147 & 55.2800 & 75.6525 & 51.77805528 & 3.8082 & 3.943783229 & 74.94462541 & 82.00651466\\
\midrule
\multicolumn{2}{l}{Spearman’s corr} & 0.95 & 0.95 & 0.56 & 0.71 & 0.50 & 0.66 & 0.90 & -0.08 & -0.21 & -0.85 & -0.89 & 0.61 & 0.83\\
\bottomrule
         \end{tabular*}
\caption{Scores on the 2023 subcorpus. Reference-based measures exclude the oracle translation itself from each source's reference set (9 references; 6 sources without any remaining reference are dropped, $n{=}301$). For TER (lower is better), we report the correlation of $-$TER, so positive values indicate agreement with manual judgments.}
\label{tab:res_2023}
\end{table*}

In terms of meaning preservation and overall translation quality, all models, except Gemini, treat high-quality human translations similarly to literal translations (see \Cref{tab:llm_as_judge_meaning} and \ref{tab:llm_as_judge_good}). Gemini and GPT  exhibit opposite tendencies in assigning the labels partial meaning preservation and no meaning preservation to both human translations and literal ones. An example of Gemini incorrectly judging a human translation is \emph{Est-ce que ce ordinateur va durer cinq ans? Obsolètement!} of the pun \emph{Will this computer last five years? Obsoletely!}. The heterographic pun relies on the same phonetic similarity \emph{absolument/obsolètement} in French as  \emph{absolutely/obsoletely} in English. While Gemini is more favorable to human translations, other models score them low (\Cref{tab:llm_as_judge_good}). Llama exhibits risk-averse behavior, assigning partial meaning preservation to both manual and machine translations and systematically judging the translations as unsuccessful.

\subsection{Machine translation measure results}

  \begin{figure}[!t]
     \centering
     \includegraphics[width=0.7\linewidth]{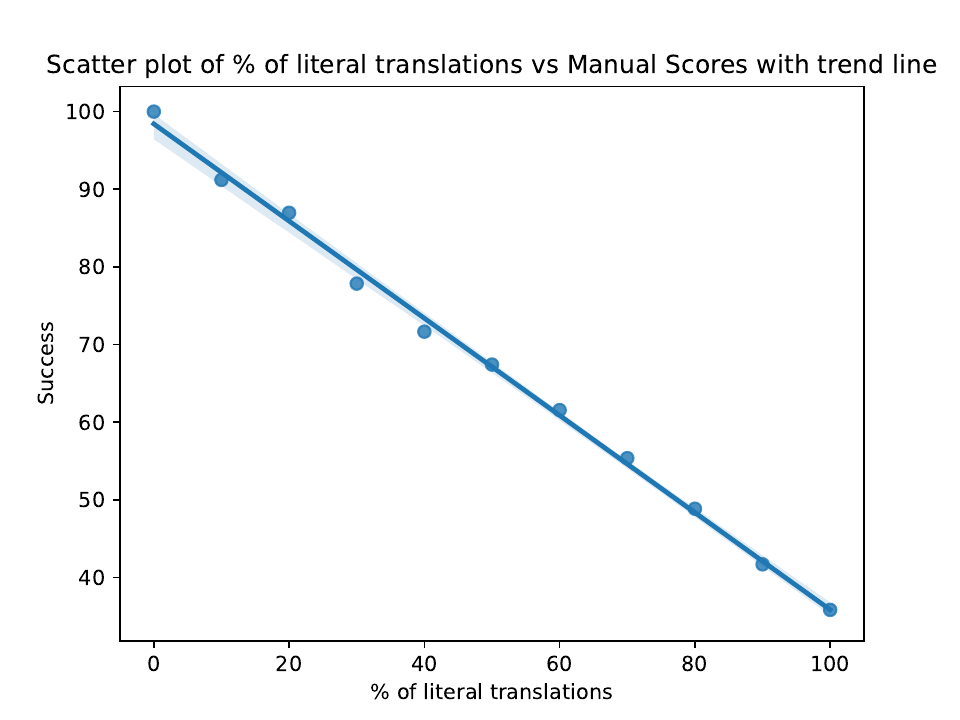}
     \caption{Literal translations (\%) vs. successful translations (\%) on the 2023 subcorpus, with regression line and 95\% confidence interval.}
     \label{fig:scatterplot_noise}
 \end{figure}

 \begin{figure}[!t]
     \centering
     \includegraphics[width=0.7\linewidth]{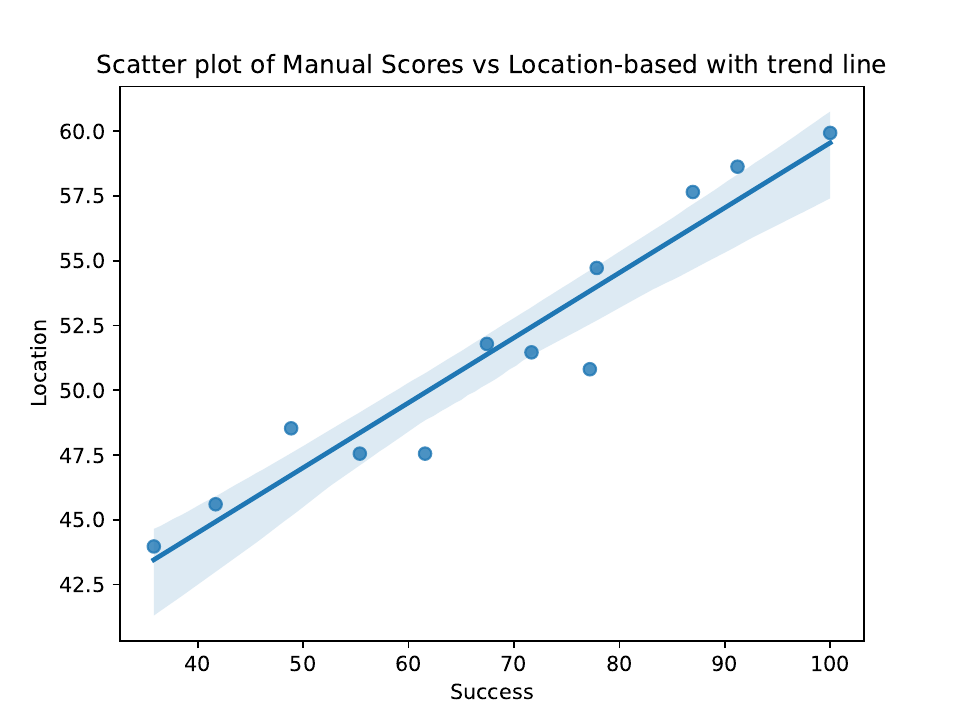}
     \caption{Scatter plot of manually attributed scores and location-based measures on the 2023 subcorpus with a regression line with a 95\% confidence interval}
     \label{fig:scatterplot2023}
 \end{figure}

\Cref{tab:res_2023} shows the percentage of successful translations by manual evaluation, as well as machine translation measure scores, on the 2023 subcorpus. As expected, increasing the proportion of literal translations in the oracle run linearly decreases the percentage of successful translations (see \Cref{fig:scatterplot_noise}). Traditional measures fail to capture this trend, whereas our location-based measure reflects it closely. The location-based measure correctly ranks the DSGT run between the oracle run and the literal translations, whereas most other measures rank it below the literal translations; BLEURT and COMETKiwi place it only marginally above them. The exception is GEMBA-MQM, which, with either backbone, assigns much higher scores to the DSGT run than to the professional translations. The Spearman's rank correlation coefficient between the manually assigned scores and the location-based measures on the 2023 subcorpus is 0.95, indicating a strong positive correlation.

Figure~\ref{fig:scatterplot2023} illustrates this relationship with a regression line and a 95\% confidence interval.
This high correlation suggests that the location-based evaluation captures key aspects of pun translation quality that closely align with expert judgments.

\begin{table*}[!t]
\small
\setlength\tabcolsep{2pt}
         \begin{tabular*}{\textwidth}{@{\extracolsep{\fill}} l*{2}{S[table-format=3.2,round-mode=places,round-precision=2]}r*
         {5}{S[table-format=3.2,round-mode=places,round-precision=2]}}
         \toprule
         {\multirow{1}{*}{run\_id}} & {\multirow{1}{*}{\%man.\(\uparrow\)}}& {\multirow{1}{*}{\%wordplay.\(\uparrow\)}} & {\multirow{1}{*}{\#Loc.\(\uparrow\)}} & {\multirow{1}{*}{\%Loc.\(\uparrow\)}} & {\multirow{1}{*}{BLEU\(\uparrow\)}} & {\multirow{1}{*}{chrF\(\uparrow\)}} & {\multirow{1}{*}{TER\(\downarrow\)}} & {\multirow{1}{*}{BERT\(\uparrow\)}}
         \\
         \midrule
dsgt\_o4\_mini\_multi\_agent\_discriminator & 79.48717948717949 & 97.43589743589743 & 332 & 14.580588493631971 & 18.9742 & 40.5208 & 80.2811 & 0.7925654053688049\\
duth\_xanthi\_argos & 41.30434782608695 & 52.17391304347826 & 305 & 13.39481774264383 & 27.4769 & 47.1599 & 66.6900 & 0.8239743709564209\\
pjmathematician\_Q25-14 & 36.36363636363637 & 38.63636363636363 & 308 & 13.526570048309178 & 28.7915 & 48.4701 & 66.6416 & 0.829602062702179\\
Skommarkhos\_Lucie\_SFT\_ARPO & 33.33333333333333 & 42.22222222222222 & 305 & 13.39481774264383 & 30.9843 & 50.7990 & 65.1142 & 0.8346604108810425\\
UvA\_finetunedNLLB-1.3B & 32.55813953488372 & 41.86046511627907 & 304 & 13.350900307422048 & 29.6585 & 49.7820 & 65.5404 & 0.8329316973686218\\
Cryptix & 30.434782608695656 & 41.30434782608695 & 296 & 12.999560825647782 & 27.4347 & 47.0442 & 67.0484 & 0.8243632316589355\\
yourteamid\_marianmt\_pun\_postedit & 30.434782608695656 & 41.30434782608695 & 296 & 12.999560825647782 & 27.4347 & 47.0442 & 67.0484 & 0.8243632316589355\\
teamX\_aug & 25.0 & 43.18181818181818 & 256 & 11.24286341677646 & 20.0403 & 40.8353 & 74.8660 & 0.8062019348144531\\
copy & 4.25531914893617 & 4.25531914893617 & 164 & 7.20 & 0.5711 & 15.2949 & 101.0893 & 0.7564066052436829\\
\midrule
Spearman's rank correlation (success) & 1.00 & 0.58 & 0.97 & 0.95 & 0.38 & 0.38 & 0.38 & 0.23\\
\midrule
Spearman's rank correlation (wordplay) & 0.58 & 1.00 & 0.46 & 0.36 & 0.04 & 0.04 & 0.04 & -0.11\\
\bottomrule
         \end{tabular*}
\caption{2025 subcorpus scores (best run per team). BERTScore is computed against the first available reference.}
\label{tab:res_2025}
\end{table*}

\Cref{tab:res_2025} presents the best-performing JOKER 2025 participant run for each team. As in 2023, the location-based measure aligns with human judgments, assigning the highest score to the best-performing run. In contrast, other measures may assign it a lower score, possibly because it is non-literal. The Spearman rank correlation between the manually assigned scores and the location-based measures on the 2025 subcorpus, across the runs, is 0.95, indicating a strong positive relationship with successful translations. 

\begin{figure}[!t]
     \centering
     \includegraphics[width=0.7\linewidth]     {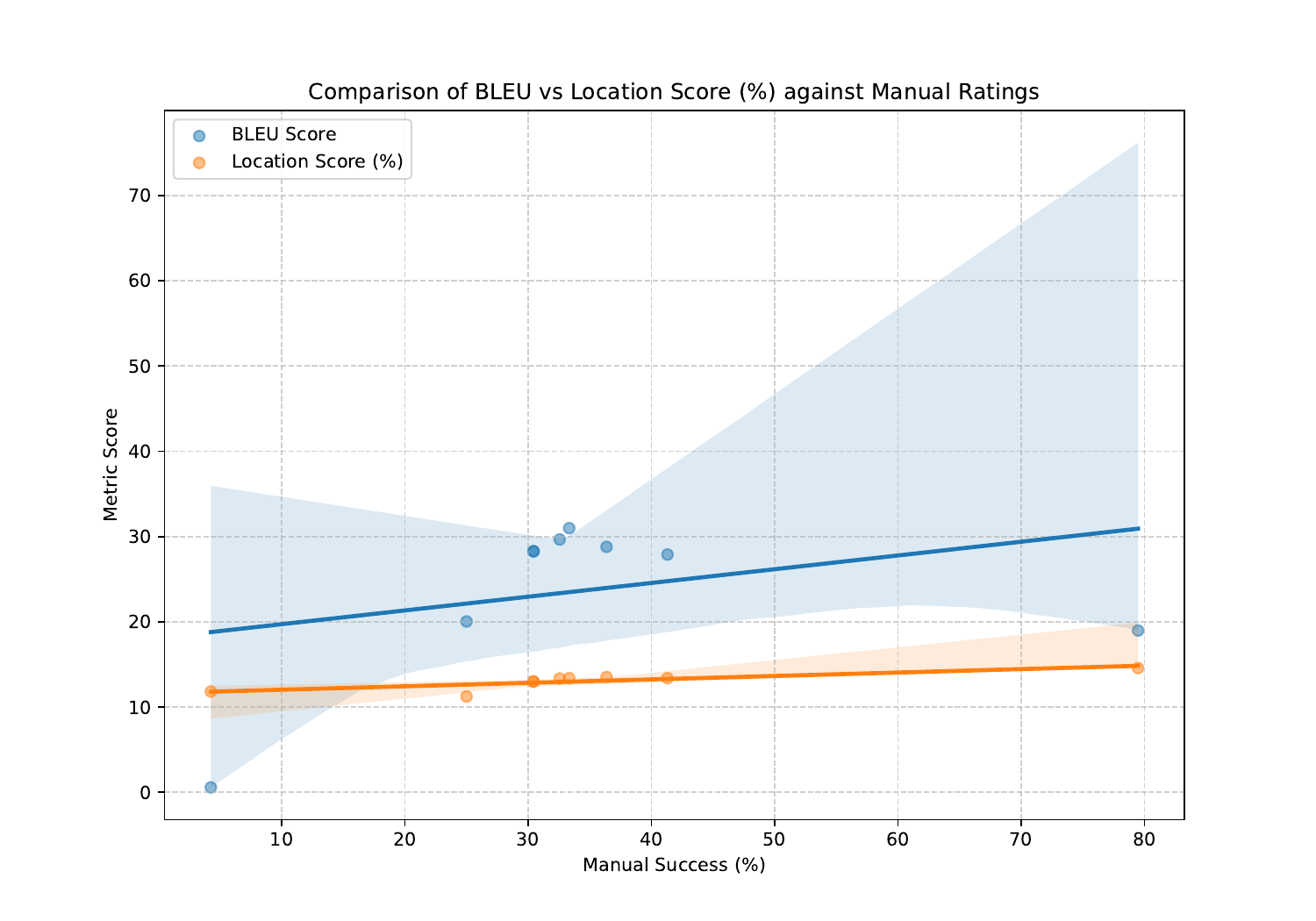}
     \caption{Scatter plot of location-based measure and BLEU against manual scores on the 2025 subcorpus with a regression line with a 95\% confidence interval}
     \label{fig:scatterplot2025}
 \end{figure}

Figure~\ref{fig:scatterplot2025} shows a clear positive relationship between the manual scores and the location-based measure on the
runs, with higher location scores generally corresponding to higher success rates. The correlation with human evaluation for the location measure on both subcorpora is substantially higher than that of traditional machine translation measures and state-of-the-art LLM-based measures. Moreover, \Cref{tab:res_2025} shows that our measure correlates much better than other measures with the presence of wordplay as annotated by humans.

\subsection{Error analysis}
We have conducted an error analysis of the location measure on the 2025 subcorpus. The false positive (FP) rate of the location measure, when evaluated against human labels for wordplay, is 5.7\%. Among them, 92\% of translations do not preserve meaning. The FP rate against successful translations is 8\%. Among them, 71.7\% do not preserve meaning, while 22.6\% show partial preservation. Only 5.7\% of FP errors preserve meaning. 

To analyze false negatives (FN), we applied the French Snowball stemmer to each word in the translations and reference wordplay locations. We found that 14\% of the location metric's false negatives, relative to successful translations, and only 10\% of its false negatives, relative to human wordplay annotations, were due to different inflected forms sharing the same stem. In most cases, this is due to a lack of references. E.g., for the translation “\textsl{Alors, comment va le business des fleurs? Oh, il est en pleine floraison}”, our reference locations are \textsl{“s'épanouit”, “fleurit”}.  Idem for “\textsl{Quand Einstein est rentré chez lui sur une locomotive, c’était un bon train de pensée}”, we have only \textsl{“fil”, “au courant”}.  Across 173 annotations, the minimum normalized edit distance between each newly annotated location and the reference locations in the corresponding manual translation averaged 0.70 (SD = 0.37). The median was 0.67, with the 25th and 75th percentiles at 0.50 and 1.00, respectively. Distances ranged from 0 to 2.40. The mean of 70\% indicates that FNs are due to missing references. 

Note that in many cases, preserving distorted spelling is essential for the wordplay to remain recognizable. For example, in “Les politiciens devraient boire plus de thé car ça leur conférerait plus d'honnête-THÉ”, “honnête-THÉ” intentionally replaces “honnêteté”. Normalizing the spelling would make the joke much harder to detect.

\section{Conclusion}
\label{sec:conclusion}
We examined the limitations of current evaluation practices for pun translation, highlighting gaps in how translation quality is assessed. We introduced a new benchmark for English–French pun translation, combining human and machine-generated translations annotated for meaning preservation and wordplay. We proposed a location-based measure that explicitly captures whether the pun-bearing span is maintained in translation, providing a complementary perspective to existing measures. It strongly correlates with human judgments, while traditional measures fail to distinguish between literal translations and those that preserve wordplay.

State-of-the-art LLMs, such as GPT-5.1 and Gemini-3-flash, often mislabel literal translations as containing wordplay, possibly due to sycophancy or the assumption that source English wordplay carries over into French. Further research is needed to investigate the underlying causes. In terms of meaning preservation and overall translation quality, both models fail to distinguish high-quality human translations from literal translations.

Although GEMBA-MQM  shows relatively high rank correlation, its strong preference for the DSGT run over the professional translations is concerning. Evaluators judged 77\% of DSGT translations successful. This proportion is roughly comparable to the 30\% of literal translations introduced into the professional translation set. Recent work shows that different LLMs repeatedly produce similar responses to open-ended tasks
\cite{Artificial_Hivemind_2026}. Both backbones may favor generated translations and undervalue valid professional alternatives. If these preferences shape translation selection or model training, readers may encounter less linguistic and creative variety over time.

We plan to develop a family of location-based metrics that account for spelling variants and jointly assess wordplay presence and meaning preservation. 
Future work may be done on other forms of wordplay and
other language pairs.

\section{Limitations}
Our experiments are restricted to two languages, only English-to-French translations.  More languages and more diverse language pairs would be of obvious value. 
These languages are diverse in terms of language typology, representing two major typological branches. English is a West Germanic language while French is a Romance language, although both are Indo-European languages. English and French have been in intense language contact for centuries, leading to some lexical borrowing and calques, but this also makes pun translation particularly relevant for this language pair. 

The source English puns are restricted to a limited number of wellerisms, such as Tom Swifty jokes, \emph{Old soldiers never die}, riddles etc. Other types of wordplay may require specialized test collections and measures. The source puns were context-free, whereas context-dependent translation may introduce additional constraints while also enabling compensatory or explanatory strategies.  While this presents an important case of wordplay, extending this to a broader set would be of clear value. Note that this task is also a linguistically important case of humorous text, and even a good use case for figurative and non-literal meaning. Hence, there is broader impact beyond the specific pun translation setting, as non-literal meaning (including proverbs and metaphors) is one of the main remaining issues in current NLP models. Our evaluation focuses on wordplay, which limits the extent to which our findings generalise to humour more broadly. Future research on other types of humor and other non-literal meanings is needed.

The current version of the location-based measure does not take into account contextual information.  In fact, the measure is relatively straightforward, and more advanced statistical or learned measures would be possible and potentially lead to higher correlations with human annotations. 
However, the current measure gives a directly interpretable score and identifies clear success and failure cases, rather than aiming for a high correlation in general. As the straightforward measure already obtains a very high correlation (95\%), there is value in a directly interpretable measure that provides direct actionable feedback on specific pun translation cases. Arguably, such a measure is still preferable over indirect indicators that may correlate even higher with laborious manual assessment. 

The location metric does not fully capture meaning preservation. On the 2025 JOKER subcorpus, its false-positive rate was 5.7\% against human wordplay labels, and 92\% of these false positives failed to preserve the source meaning. Against human judgments of translation success, the false-positive rate was 8\%. Of these false positives, 71.7\% lost the source meaning and 22.6\% preserved it only partially. Future work will test whether combining wordplay detection with an explicit assessment of meaning preservation improves the identification of successful translations.

The automatic runs were initially annotated by a single annotator. As the main references come from human professional translators, we only conducted a secondary validation of their references, revealing a very high agreement of 98\%.  One cause of the very high agreement is the quality of human professional translations on which our references are based.
When annotating low-quality system predictions, we may encounter more edge cases and observe slightly lower agreement between different annotators. 

\section*{Acknowledgments}
Liana Ermakova is supported by a government grant managed by the National Research Agency under the program Investissements d'avenir integrated into France 2030, with the Reference ANR-19-GURE-0001. We thank all the translators, students, and members of the team at the University of Brest’s master’s program in translation who contributed to building the dataset.

Jaap Kamps is supported by the Netherlands Organization for Scientific Research (NWO NWA \# 1518.22.105), the University of Amsterdam (AI4FinTech program), and ICAI (AI for Open Government Lab). 
Views expressed in this paper are not necessarily shared or endorsed by those funding the research.

\paragraph{\sl Disclosure of Interests.}
The authors have no competing interests to declare that are relevant to the content of this article.

\paragraph{\sl Disclosure of Generative AI use.}
During the preparation of this work, the authors used ChatGPT and Grammarly 
for spelling checks and paraphrasing.
Further, the authors used Gemini in order to generate images.
After using these tools\slash services, the authors reviewed and edited the content as needed and take full responsibility for the publication’s content.

\bibliography{references}

\appendix

\crefname{section}{appendix}{appendices}
\Crefname{section}{Appendix}{Appendices}
\crefalias{section}{appendix}

\section{Manual Translation and Annotation}
\label{appendix:agreement}
The dataset includes French translations produced by professional translators, native French speakers, and automatic runs submitted to the JOKER track in 2023 and 2025. We sourced manually created translations from a human contest aimed at translating puns from English into French while both preserving meaning and including a form of wordplay \cite[similar to][]{DBLP:conf/sigir/ErmakovaBJM23}. 

We further extended this corpus by adding new translations produced by master’s students in translation, all of whom are native French speakers.
Interns who were also master’s students in translation and native French speakers completed the annotation. 
Translators and annotators consented to the use of the data for research purposes. Translation interns were compensated in accordance with the university’s internship payment regulations. 

\begin{figure*}[!ht]
    \centering
    \includegraphics[width=\linewidth]{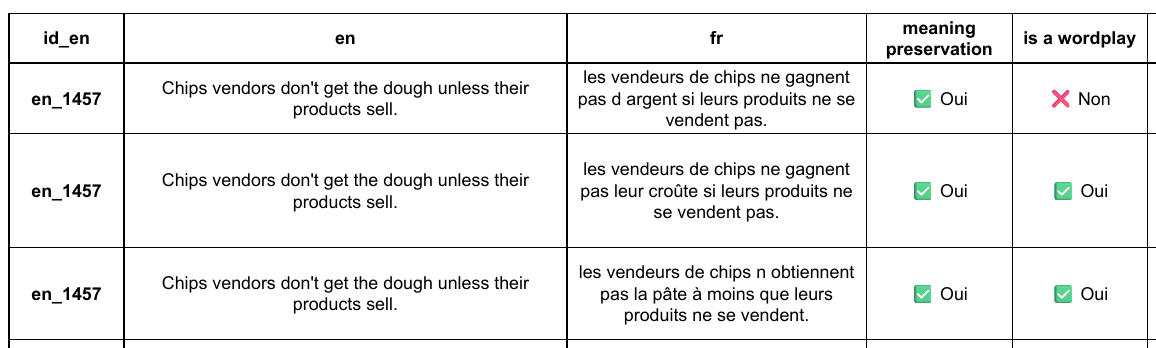}
    \caption{Run evaluation interface}
    \label{fig:interface_eval}
\end{figure*}

JOKER runs were evaluated for meaning preservation and the presence of wordplay in the French translation, which may take forms other than a pun (e.g., alliteration, spoonerisms/contrepèteries, etc.). The interface is presented in \Cref{fig:interface_eval}.

\begin{figure*}[!ht]
    \centering
    \includegraphics[width=\linewidth]{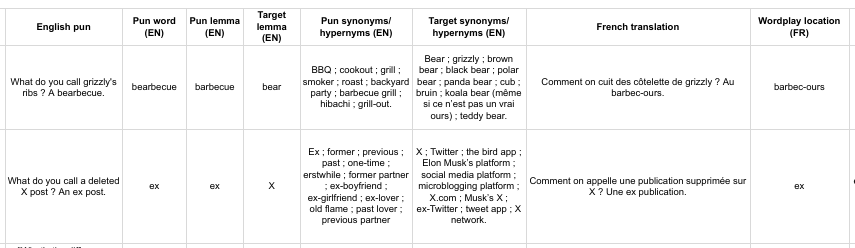}
    \caption{Pun location annotation interface}
    \label{fig:interface_location}
\end{figure*}

We conducted the location annotation in Google Sheets. Annotators were provided with French translations of the English puns, along with semantic annotations from \citet{miller-etal-2017-semeval}, later adapted by \citet{DBLP:conf/sigir/ErmakovaBJM23}. \Cref {fig:interface_location} shows a screenshot of the pun location annotation interface.

When the second annotator added location spans to the 2,087 previously identified instances of wordplay, 38 cases were reclassified as non-wordplay. This corresponds to a simple agreement rate of 98.18\%. Among the 38 disagreements, 35 translations were identified as literal translations without wordplay, one French translation was judged nonsensical, one case involved an incomplete translation, and one resulted from a transcription error.\footnote{This only concerns false positives, hence would correspond to an interrater agreement is estimated at 96.36\%.}
    
We automatically verified the presence of each location in the French translation using exact string matching. For the 611 cases in which a location was not detected, we conducted a manual review. Among these, 44 cases were due solely to apostrophe variation, where the source used ' and the French translation used ’, or vice versa. In 148 cases, the translated wordplay did not constitute puns in the strict sense of double meaning, but instead relied on other phonetic phenomena such as homophony. For example, \emph{Pourquoi le ver de terre passe-t-il sa soirée au bar ? Pour boire un verre}, the annotated location was \emph{verre | ver de terre}. Because these two expressions are separated in the sentence, our automatic string-matching procedure incorrectly flagged them as errors. In 2 additional cases, annotators provided an interpretation instead of a location span. The remaining discrepancies were orthographic errors or differences in word form; for example, annotators occasionally provided an infinitive form as the location, whereas the French translation contained a conjugated form of the verb.

\section{LLM-as-a-judge Parameters}
\label{appendix:llm_param}
For all Gemini 3 models, Google recommends keeping the temperature at the default 1.0 because lowering it may lead to unexpected behavior, such as looping or degraded performance\footnote{\url{https://ai.google.dev/gemini-api/docs/gemini-3?hl=en}}. Thus, we run Gemini\footnote{\url{https://ai.google.dev/gemini-api/docs/models/gemini-3-flash-preview}} with the default parameters. We run GPT-5.1\footnote{\url{https://developers.openai.com/api/docs/models/gpt-5.1}}
with \emph{temperature=0, top\_p=0} (\emph{top\_k} and \emph{seed} are not available). We run Qwen\footnote{\url{https://huggingface.co/Qwen/Qwen3-4B-Instruct-2507}} with \emph{max\_new\_tokens=1} and \emph{do\_sample=False} (in this case, \emph{temperature, top\_p}, and \emph{top\_k} are ignored, \emph{seed} is not available). Llama\footnote{\url{https://huggingface.co/meta-llama/Llama-3.1-8B-Instruct}} was run with \emph{do\_sample=False, temperature=0, top\_p=0, top\_k=1, max\_new\_tokens=1}. 

\section{LLM-as-a-judge Prompts}
\label{appendix:llm_prompts}

In the first experiment, we directly asked the models to evaluate translation quality using the following prompt:
\begin{Verbatim}[breaklines,fontsize=\footnotesize,gobble=4]
    f"You are an expert translator from English into French. Is the following text in French a good translation of the source English text?
    English source: {en}
    French translation: {fr}
    Answer only Yes or No, nothing else."
\end{Verbatim}

In the second experiment, we mimic our human evaluation by independently assessing two aspects of the pun translation: meaning preservation and the presence of wordplay in the French version. A translation is considered successful if it preserves the meaning and contains wordplay.
\begin{Verbatim}[breaklines,fontsize=\footnotesize,gobble=4]
    f"You are an expert translator from English into French. Does the following translation in French preserve the meaning of the source English text?
    English source: {en}
    French translation: {fr}
    Answer only Yes, No, or Partially, nothing else."
\end{Verbatim}
\begin{Verbatim}[breaklines,fontsize=\footnotesize,gobble=4]
    f"You are an expert in wordplay in French. Is the following text a wordplay?
    Text: {fr}
    Answer only Yes or No, nothing else."
\end{Verbatim}

\section{LLM-as-a-judge Post-processing}
\label{appendix:postprocessing}
Although we asked models to answer only Yes or No, a large number of answers were in French or contained other tokens. We normalized the outputs by translating them into English and extracting only Yes/No answers for wordplay detection and \emph{good translations}. For meaning preservation, we homogenized the outputs into Yes/No/Partial. The outputs were truncated after these tokens. We assumed that -1 means No. Llama produced the output O in 30\% of cases for wordplay detection. In this case, we assumed that it was a first token of the French Oui (Yes), so we replaced these values by Yes. Note that Llama produced occasional errors like \emph{The}. 

\section{Examples of the effect of prompt framing on LLM judgments about wordplay}
\label{app:sycophancy_judge}
Our previous work showed that LLMs can exhibit sycophantic behaviour when asked about wordplay, endorsing opposing interpretations of the same text depending on whether the prompt presents it as a pun or a non-pun \cite{DBLP:conf/chiir/ErmakovaFK26}. \Cref{fig:sycophancy_ex} provides examples of the effect of prompt framing on Gemini answers prompted to explain why "I think my learning curve has turned into a circle" is a pun and why it is not a pun. These examples are of interest from a sycophancy-analysis perspective, as LLMs' inability to model non-literal meaning inherently leads to sycophantic traits dominating their responses.  While one of the responses makes perfect sense, flipping the prompt's stance yields the opposite response.  While the second, negative prompt may seem unnatural, the first prompt takes a very similar positive stance.   This observation also demonstrates the value of our resource and the enormous effort to collect true wordplay translations from professional translators. 

\begin{figure}[!t]
  \footnotesize
\newcommand{\myfont}{\fontsize{8}{8}\selectfont}
\setlength\tabcolsep{0.008\linewidth}
\begin{tabular*}{\linewidth}{|>{\myfont\sl}p{0.98\linewidth}|}
\multicolumn{1}{c}{\myfont\bf Positive assumption in prompt}\\
\hline
  Explain the following pun: "I think my learning curve has turned into a circle."\\
  "This pun plays on two related but contrasting ideas: 1. **Learning Curve:** * **Literal Meaning:** In mathematics or statistics, a learning curve is a graph that shows how a person's performance or knowledge improves over time or with practice. It typically starts low and then goes up, indicating progress and acquisition of skill/knowledge. * **Common Usage:** Metaphorically, "learning curve" refers to the process of acquiring a new skill or understanding new information. A "steep learning curve" means it's challenging at first but you learn quickly, while a "shallow learning curve" means it takes a long time to learn. The key idea is **progress** – you are moving forward, gaining understanding. 2. **Turned into a Circle:** * A circle, unlike a curve that moves progressively in one direction, is a shape that **returns to its starting point** continually. * When applied to a process, "going in circles" or a "vicious circle" means **repetition, lack of progress, being stuck, or relearning the same things over and over without ever truly advancing.** **The Pun's Meaning:** The humor comes from the ironic contrast. The speaker expects their learning process (their "learning curve") to show forward movement and progress. Instead, they feel like they are no longer progressing; they are just **repeating the same steps, encountering the same difficulties, or forgetting what they've learned, effectively ending up back where they started.** In short, the pun means: **"I feel like I'm stuck, not making any real progress in my learning, and keep going over the same material or making the same mistakes repeatedly."**"\\
  \hline
\multicolumn{1}{c}{\myfont ~}\\[-1ex]
\multicolumn{1}{c}{\myfont\bf Negative assumption in prompt}\\
\hline
  Explain why the following text is NOT a pun: "I think my learning curve has turned into a circle."\\
  "The phrase "I think my learning curve has turned into a circle" is **NOT a pun**. Here's why: 1. **What is a Pun?** A pun is a form of wordplay that exploits multiple meanings of a word, or of words that sound alike but have different meanings, for an intended humorous or rhetorical effect. It relies on **linguistic ambiguity** based on sound or spelling. 2. **Why the phrase isn't a pun:** * There are no words that sound alike but have different meanings (homophones). "Curve" is "curve," "circle" is "circle." * There's no single word being used in two different senses simultaneously. * The humor or insight doesn't come from exploiting the sound or spelling of a word. 3. **What the phrase *is*:** It's an **extended metaphor** or a **figurative expression** (specifically, an idiom or a play on an idiom). * A "learning curve" is an idiom that refers to the rate at which someone acquires a new skill or knowledge. Typically, a curve suggests progress over time (steep at first, then flattening out). * By saying it "turned into a circle," the speaker is using the geometric shape of a circle to metaphorically represent **repetition, stagnation, going nowhere, or returning to the beginning without having made progress.** The cleverness or humor in the phrase comes from the **irony** and the **visual metaphor** of a line of progress becoming a closed loop, rather than from wordplay. It's a vivid way to express frustration about a lack of advancement."\\
  \hline
  \end{tabular*}
  \caption{Gemini responses to the \emph{exact same sentence} when asked why it is a pun (top) or not a pun (bottom).
  }
  \label{fig:sycophancy_ex}
\end{figure}

\section{Resources}
\label{appendix:resource}

The GitHub repository is available at \url{https://github.com/joker-project/puns-lost-in-translation}. It provides a subset of the English–French pun translation corpus, results from the LLM-as-a-judge evaluation, implementations of the proposed location-based measure and the baseline measures, and code illustrating their use. We have reserved the remaining data for our follow-up studies and for future editions of the CLEF JOKER shared task, see \url{https://www.joker-project.com}. 
Researchers may request access to additional data by contacting the authors.

\end{document}